\documentclass[runningheads]{llncs}
\pdfoutput=1

\usepackage[utf8]{inputenc}
\usepackage[british]{babel}
\usepackage{amsmath,amssymb}
\usepackage{setspace}
\usepackage{relsize}
\usepackage{tikz,pgf}
\usetikzlibrary{backgrounds,shapes,decorations,snakes}
\usepackage{url}
\usepackage{color}
\usepackage[pdftex,colorlinks=true,linkcolor=black,citecolor=black,menucolor=black,urlcolor=black]{hyperref}
\usepackage[font=it]{caption}
\usepackage{color}
\usepackage{stmaryrd}
\usepackage{listings}  

\newcounter{cexample}
\newcounter{csection}

\ifdefined\implies
 \renewcommand{\implies}{\supset}
\else
 \newcommand{\implies}{\supset}
\fi

\newcommand{\tvt}{\mathbf{t}}
\newcommand{\tvf}{\mathbf{f}}
\newcommand{\tvu}{\mathbf{u}}

\newcommand{\lin}{\mathit{in}}
\newcommand{\lout}{\mathit{out}}

\newcommand{\NP}{\ensuremath{\mathbf{NP}}}

\newcommand{\parents}[1]{\mathit{par}(#1)}

\newcommand{\define}[1]{\emph{#1}}
\newcommand{\set}[1]{\left\{#1\right\}}

\newcommand{\guard}{\ \middle\vert\ }

\newcommand{\ileq}{\leq_i}
\newcommand{\ilt}{<_i}

\newcommand{\la}{\ensuremath{\leftarrow}}

\newenvironment{slisting}{%
\smaller\ttfamily%
\begin{tabbing}
\hspace*{1cm} \= \kill}
{%
\end{tabbing}
}

\definecolor{gray}{gray}{0.4}

\lstdefinelanguage{ASPenc}
{
  morekeywords={s,l,ci,co},
  morecomment=[l]{\%}
}

\newcommand{\presection}{}
\newcommand{\postsection}{}
\newcommand{\presubsection}{}
\newcommand{\postsubsection}{}

\begin{document}
\titlerunning{The DIAMOND System for Argumentation: Preliminary Report}
\authorrunning{S. Ellmauthaler and H. Stra\ss}
\setcounter{page}{97}
\title{The DIAMOND System for Argumentation: Preliminary Report\thanks{This research has been supported by DFG projects BR~1817/7-1 and FOR~1513.}}
\author{Stefan Ellmauthaler and Hannes Strass}
\institute{Computer Science Institute, Leipzig University}
\maketitle

\begin{abstract}
  Abstract dialectical frameworks (ADFs) are a powerful generalisation of Dung's abstract argumentation frameworks.
  In this paper we present an answer set programming based software system, called DIAMOND (DIAlectical MOdels eNcoDing). 
  It translates ADFs into answer set programs whose stable models correspond to models of the ADF with respect to several semantics (i.e.\ admissible, complete, stable, grounded).
\end{abstract}


\presection
\section{Introduction}
\label{sec:introduction}
\postsection

Formal argumentation has established itself as a vibrant subfield of artificial intelligence, contributing to such diverse topics as legal decision making and multi-agent interactions.
A particular, well-known formalism to model argumentation scenarios are Dung's abstract argumentation frameworks~\cite{dung95acceptability}.
In that model, arguments are treated as abstract atomic entities.
The only information given about them is a binary relation expressing that one argument attacks another.

A criticism often advanced against Dung frameworks is their restricted expressive capability of allowing only attacks between arguments.
This leads to quite a number of extensions of Dung AFs, for example with
attacks from sets of arguments~\cite{NielsenP06},
attacks on attacks~\cite{Modgil09} and
meta-argumentation~\cite{DBLP:journals/sLogica/BoellaGTV09}.
Unifying these and other extensions to AFs, Brewka and Woltran~\cite{brewka-woltran10adfs} proposed a general model, abstract dialectical frameworks (ADFs).
In ADFs, attack, support, joint support, combined attacks and many more relations between arguments (called statements there) can be expressed, while the whole formalism stays on the same abstraction level as Dung's.

In this paper we present the DIAMOND software system that computes models of ADFs with respect to several different semantics.
The name DIAMOND abbreviates ``DIAlectical MOdels eNcoDing'' and hints at the fact that DIAMOND is built on top of the state of the art in answer set programming:
abstract dialectical frameworks are encoded into logic programs, and an answer set solver is used to compute the models of the ADF.
By providing an expressive argumentation formalism with an implementation, we pave the way for practical applications of ADFs in scenarios where dialectical aspects are of interest, for example in group decision making.

The paper proceeds as follows.
We first introduce the necessary background in abstract dialectical frameworks and answer set programming.
We then present the DIAMOND system -- how ADFs are represented there, and how the ADF semantics are encoded into answer set programs.
We conclude with a contrasting discussion of the most significant related work.

\presection
\section{Background}
\label{sec:background}
\postsection

An abstract dialectical framework (ADF)~\cite{brewka-woltran10adfs} is a directed graph whose
nodes represent statements or positions that can be accepted or
not. The links represent dependencies: the status of a node $s$ only
depends on the status of its parents (denoted $\parents{s}$), that
is, the nodes with a direct link to $s$. In addition, each node $s$
has an associated acceptance condition $C_s$ specifying the exact
conditions under which $s$ is accepted. $C_s$ is a function
assigning to each subset of $\parents{s}$ one of the truth values $\tvt$,
$\tvf$.
Intuitively, if for some $R \subseteq \parents{s}$ we have $C_s(R) = \tvt$, then $s$ will be accepted
provided the nodes in $R$ are accepted and those in $\parents{s} \setminus R$ are not accepted.

\begin{definition}\label{def:ADF}
 An \emph{abstract dialectical framework} is a tuple $D = (S, L, C)$ where
\begin{itemize}
\item $S$ is a set of statements (positions, nodes),
\item $L \subseteq S \times S$ is a set of links,
\item $C = \{C_s\}_{s \in S}$ is a set of total functions $C_s : 2^{\parents{s}}\rightarrow \{\tvt,
\tvf\}$.
\end{itemize}
\end{definition}
In many cases it is convenient to represent the acceptance condition of a statement $s$ by a propositional formula $\varphi_s$, as is done in our running example. 
\begin{example}
  \label{exa:ADF}
  Consider the ADF \mbox{$D=(S,L,C)$} with a support cycle and one attack relation:
  \mbox{$S=\set{a,b,c}, L=\set{(a,b),(b,a),(b,c)}, \varphi_a=b, \varphi_b=a, \varphi_c = \neg b$}.
  This ADF can also be represented as a graph, where the nodes are statements and the relations between them are directed edges. The boxes below each node are the acceptance conditions for the particular statement.
  
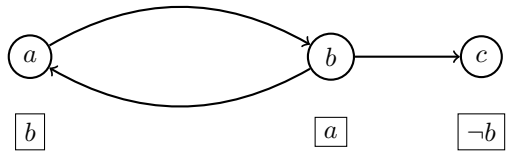
\begin{center}\begin{tikzpicture}[->,thick,circle]
  \node (a) at (0,0) [draw] {$a$};
  \node (b) at (4,0) [draw] {$b$};
  \node (c) at (6,0) [draw] {$c$};

  \path (a) edge [bend left] (b);
  \path (b) edge [bend left] (a);
  \draw (b) -- (c);

  \node (a1) at (0,-1)[draw,thin,rectangle]{$b$};
  \node (b1) at (4,-1)[draw,thin,rectangle]{$a$};
  \node (c1) at (6,-1)[draw,thin,rectangle]{$\lnot b$};
\end{tikzpicture}

\end{center}
\end{example}

In recent work~\cite{brewka13adfs}, we redefined several standard ADF semantics and defined additional ones.
In this paper, we use these revised definitions, which are based on three-valued logic.\footnote{For further details on those newly introduced semantics we refer the interested reader to Brewka et al.~\cite{brewka13adfs}.}
The three truth values true ($\tvt$), false ($\tvf$) and unknown ($\tvu$) are partially ordered by $\ileq$ according to their information content:
we have \mbox{$\tvu\ilt\tvt$} and \mbox{$\tvu\ilt\tvf$} and no other pair in $\ilt$, which intuitively means that the classical truth values contain more information than the truth value unknown.
On the set of truth values, we define a meet operation, \emph{consensus}, which assigns
\mbox{$\tvt\sqcap\tvt = \tvt$}, \mbox{$\tvf\sqcap\tvf = \tvf$}, and returns $\tvu$ otherwise.
The information ordering $\ileq$ extends in a straightforward way to valuations $v_1,v_2$ over $S$ in that
\mbox{$v_1 \ileq v_2$} iff \mbox{$v_1(s) \ileq v_2(s)$} for all \mbox{$s\in S$}.
Obviously, a three-valued interpretation $v$ is two-valued if all statements are mapped to either true or false.
For a three-valued interpretation $v$, we say that a two-valued interpretation $w$ \define{extends} $v$ iff \mbox{$v \ileq w$}.
We denote by $[v]_2$ the set of all two-valued interpretations that extend~$v$.
A three-valued interpretation $E_v$ has an \define{associated extension} 
\mbox{$E_v = \set{ s\in S \guard v(s)=\tvt }$}.

Brewka and Woltran~\cite{brewka-woltran10adfs} defined an operator $\Gamma_D$ over three-valued interpretations.
For each statement $s$, the operator returns the consensus truth value for its acceptance formula $\varphi_s$, where the consensus takes into account all possible two-valued interpretations $w$ that extend the input valuation $v$.

\begin{definition}
  \label{def:semantics}
  Let $D$ be an ADF and $v$ be a three-valued interpretation. 
  Then the interpretation $\Gamma_D(v)$ is given by
  \mbox{$s \mapsto \bigsqcap \set{ w(\varphi_s) \guard w\in[v]_2 }$}.
  Furthermore $v$ is
  \define{admissible} iff $v\ileq\Gamma_D(v)$;
  \define{complete} iff $\Gamma_D(v) = v$, that is, $v$ is a fixpoint of $\Gamma_D$;
  \define{grounded} iff $v$ is the $\ileq$-least fixpoint of $\Gamma_D$.

  A two-valued interpretation $v$ is a 
  \define{model of $D$} iff $\Gamma_D(v)=v$;
  it is a \define{stable model} of \mbox{$D = (S, L, C)$} iff
  $v$ is a model of $D$ and
  $E_v$ equals the grounded extension of the reduced ADF
  \mbox{$D^v = (E_v, L^v, C^v)$}, where
  \mbox{$L^v = L \cap (E_v\times E_v)$} and
  for \mbox{$s \in E_v$} we set
  \mbox{$\varphi^v_s = \varphi_s[r/\tvf\,:\,v(r)=\tvf]$}.
\end{definition}


\begin{example}
  We will now show the models with respect to the different semantics for the ADF introduced in Example~\ref{exa:ADF}.
  For readability, we write interpretations $v$ 
  as sets of literals
  \mbox{$L_v = \set{ s\in S \guard v(s)=\tvt } \cup \set{ \neg s \guard s\in S, v(s)=\tvf }$}.
  There are
  \begin{itemize}
  \item five admissible interpretations: 
    $\emptyset$,
    $\{a,b\}$,
    $\{a,b,\lnot c\}$
    $\{\lnot a,\lnot b,c\}$, 
    $\{\lnot a,\lnot b\}$,
  \item three complete models:
    $\{\lnot a,\lnot b,c\}$,
    $\emptyset$,
    $\{a,b,\lnot c\}$;
    of which $\emptyset$ is grounded;
  \item two models:
    $\{a,b,\neg c\}$, $\{\neg a, \neg b, c\}$,
    of which one is stable:
    $\{\neg a, \neg b, c\}$
  \end{itemize}
\end{example}


Brewka et al.~\cite{brewka13adfs} also defined an approach to handle preferences in ADFs.
The approach generalises the one for AFs from Amgoud and Cayrol~\cite{AmgoudC98}.
Since DIAMOND also implements this treatment of preferences, we recall it here.
For this approach, the links are restricted to links that are attacking or supporting.

\begin{definition}
A prioritised ADF (PADF) is a tuple $P = (S, L^+, L^-, >)$ where
$S$ is the set of nodes, $L^+$ and $L^-$ are subsets of $S \times
S$, the supporting and attacking links, and $>$ is a strict partial
order (irreflexive, transitive, antisymmetric) on $S$ representing preferences
among the nodes.
\end{definition}

Here \mbox{$(a,b) \in \; >$} (alternatively: \mbox{$a > b$}) expresses that $a$ is preferred to $b$.
The semantics of prioritised ADFs is given by a translation to standard ADFs:
$P$ translates to \mbox{$(S, L^+ \cup L^-, C)$},
where for each statement $s \in S$ the acceptance condition $C_s$ is
defined as:
$C_s(M) = \tvt$ iff for each $a \in M$ such that $(a,s) \in L^-$ and
not $s > a$ we have: for some $b \in M$, $(b,s) \in
L^+$ and $b > a$.
Intuitively, an attacker does not succeed if the attacked node is
more preferred or if there is a more preferred supporting node.

\presubsection
\subsection{Answer Set Programming}
\label{sec:asp}
\postsubsection

A \define{propositional normal logic program} $\Pi$ is a set of finite rules $r$ over a set of ground atoms $\mathcal{A}$. A rule $r$ is of the form
\mbox{$\alpha \leftarrow \beta_1,\ldots,\beta_m, \text{not }\beta_{m+1}\ldots, \text{not }\beta_n$},
where $\alpha\in\mathcal{A}$, $\beta_i\in\mathcal{A}$ are ground atoms and $m\leq n\leq 0$. 
Each rule consists of a \define{body} $B(r)=\{\beta_1,\ldots,\beta_m, \text{not }\beta_{m+1}\ldots, \text{not }\beta_n\}$  and a \define{head} $H(r)=\{\alpha\}$, divided by the $\leftarrow$-symbol. 
We will split up the body into two parts, $B(r)=B^+(r)\cup B^-(r)$, where $B^+(r)=\{\beta_1,\ldots,\beta_m\}$ and $B^-(r)=\{\text{not }\beta_{m+1}\ldots, \text{not }\beta_n\}$. A rule $r$ is said to be positive if $B^-(r)=\emptyset$ and a program $\Pi$ is positive if every rule $r\in\Pi$ is positive. 
For a positive program $\Pi$, its immediate consequence operator $T_\Pi$ is defined for $S\subseteq\mathcal{A}$ by $T_\Pi(S) = \set{ H(r)\in\mathcal{A} \guard r\in \Pi, B^+(r)\subseteq S }$.
A set $A\subseteq\mathcal{A}$ of ground atoms is a \define{minimal model} of a positive propositional logic program $\Pi$ iff $A$ is the least fixpoint of $T_\Pi$.
To allow rules with negative body atoms, Gelfond and Lifschitz~\cite{Gelfond1988} proposed the stable model semantics (also called answer set semantics).
\begin{definition}
  Let $A\subseteq\mathcal{A}$ be a set of ground atoms. $A$ is a \define{stable model} for the propositional normal logic program $\Pi$ iff $A$ is the minimal model of the reduced program $\Pi^A$, where
  \mbox{$\Pi^A = \{H(r)\leftarrow B^+(r)\mid r\in\Pi, B^-(r)\cap A =\emptyset\}$}.
\end{definition}
We use {\em clasp} from the Potsdam Answer Set Solving Collection {\em Potassco}\footnote{Available at \url{http://potassco.sourceforge.net}}~\cite{Gebser2011} as the back-end answer set solver for our software system.
Potassco allows us to use an enriched input language where in addition to the above pictured propositional logic programs we can use first-order variables and predicates.
Ground atoms are generally written in lower case while variables are represented with upper case characters. Additionally {\em Potassco} offers features like aggregates, cardinality constraints, choice rules and conditional literals. For further details we refer to the recent book by Gebser et al.~\cite{Gebser2012}.

\presection
\section{DIAMOND}
\label{sec:diamond}
\postsection

Our software system DIAMOND is a collection of answer set programming encodings and tools to compute the various models with respect to the semantics for a given ADF. The different encodings are designed around the \emph{Potsdam Answer Set Solving Collection (Potassco)}~\cite{Gebser2011} and the additional provided tools utilise clasp as solver, too. Note that the encodings for DIAMOND are built in a modular way. To compute the models of an ADF with respect to a semantics, different modules need to be grounded together to get the desired behaviour. 

DIAMOND is available for download and experimentation at the web page \url{http://www.informatik.uni-leipzig.de/~ellmau/diamond}. \sloppy
There we also provide further documentation on its usage.
In short, DIAMOND is a Python-script,\footnote{Python is available at \url{http://www.python.org}.} which can be invoked via the command line. Different switches are used to designate the desired semantics, and the input file is given as a file name or via the standard input. The options for the command line are as follows:
\newpage
\begin{slisting}
usage: diamond.py [-h] [-cf] [-m] [-sm] [-g] [-c] [-a] \\ 
\ \ \ \ \ \ \ [--transform\_pform | --transform\_prio] [-all] [--version] instance\\

positional arguments:\\
\ \ \ instance\ \ \ \ \ \ \ \ \ \ \ \ \ \ File name of the ADF instance\\

optional arguments:\\
\ \ \   -h, --help\ \ \ \ \ \ \ \ \ \ \ \ show this help message and exit\\
\ \ \   -cf, --conflict-free\ \ compute the conflict free sets\\
\ \ \   -m, --model\ \ \ \ \ \ \ \ \ \ \ compute the two-valued models\\
\ \ \   -sm, --stablemodel\ \ \ \ compute the stable models\\
\ \ \   -g, --grounded\ \ \ \ \ \ \ \ compute the grounded model\\
\ \ \   -c, --complete\ \ \ \ \ \ \ \ compute the complete models\\
\ \ \   -a, --admissible\ \ \ \ \ \ compute the admissible models\\
\ \ \   --transform\_pform\ \ \ \ \ transform a propositional formula ADF before the computation\\
\ \ \   --transform\_prio\ \ \ \ \ \ transform a prioritized ADF before the computation\\
\ \ \   -all, --all\ \ \ \ \ \ \ \ \ \ \ compute all sets and models\\
\ \ \   --version\ \ \ \ \ \ \ \ \ \ \ \ \ prints the current version
\end{slisting}
We next describe how specific ADF instances are represented in DIAMOND.

\presubsection
\subsection{Instance Representation}
\label{sec:instancerepresentation}
\postsubsection

In order to represent an ADF for DIAMOND its acceptance conditions need to be in the functional representation as given in Definition~\ref{def:ADF}. 
The statements of an ADF are declared by the predicate {\tt s}, and the links are represented by the binary predicate {\tt l}, such that {\tt l(b,a)} reflects that there is a link from $b$ to $a$. 
The acceptance condition is modelled via the unary and tertiary predicates {\tt ci} and {\tt co}. 
Intuitively {\tt ci} (resp.\ {\tt co}) identifies the parents which need to be accepted, such that the acceptance condition maps to true (i.e.\ \emph{in}) (resp.\ false (i.e.\ \emph{out})). 
To achieve a flat representation of each set of parent statements, we use an arbitrary third term in the predicate to identify them. 
To express what happens to a statement when none of the parents is accepted we use the unary versions of {\tt ci} and {\tt co}.
Here is the DIAMOND representation of Example~\ref{exa:ADF}\/:
\begin{slisting}
    \> s(a). s(b). s(c). l(b,a). l(a,b). l(b,c). \\
    \> co(a). ci(a,1,b).  co(b). ci(b,1,a).  ci(c). co(c,1,b).     
\end{slisting}
The first line declares the statements and links.
The second line expresses the acceptance conditions:
statement $a$ is $\lout$ if $b$ is $\lout$ and $\lin$ if $b$ is;
likewise $b$ gets the same status as $a$;
statement $c$ is $\lin$ if $b$ is $\lout$, and $c$ is $\lout$ if $b$ is $\lin$.

As a part of the DIAMOND software bundle, we also provide an ECL$^i$PS$^e$ Prolog\footnote{ECL$^i$PS$^e$ is available at \url{http://eclipseclp.org/}.}~\cite{SchimpfS2010} program that transforms acceptance functions given as formulas into the functional representation used by DIAMOND.

We have chosen this functional representation of acceptance conditions for pragmatic reasons.
An alternative would have been to represent acceptance conditions by propositional formulas.
In this case, computing a single step of the operator would entail solving several \NP{}-hard problems.
The standard way to solve these is the use of saturation~\cite{EiterG95}, which however causes undesired side-effects when employed together with meta-ASP~\cite{GebserKS2011}.
Furthermore, other ADF semantics (e.g.\ preferred) utilise concepts like $\subseteq$-minimality, which also require the use of meta-argumentation.
We plan to extend DIAMOND to further semantics and therefore chose the functional representation of acceptance conditions to forestall potential implementation issues.

Due to compatibility considerations, it is possible for DIAMOND to understand the propositional formula representation as well as a PADF. The propositional formula representation uses the unary predicate {\tt statement} to identify statements. The binary predicate {\tt ac(s,$\phi$)} associates to each statement {\tt s} one formula $\phi$. Each formula {\tt $\phi$} is constructed in the usual inductive way, where atomic formulae are other statements and the truth constants (i.e.\ {\tt c(v)} and {\tt c(f)}) and the operators are written as functions. The allowed operators are {\tt neg, and, or, imp,} and {\tt iff} for their respective logical operators. To describe a PADF, we use the unary predicate {\tt s} to describe the set of statements. In addition the support (i.e.\ $L^+$) and attack (i.e.\ $L^-$) links are represented by the binary predicates {\tt lp} and {\tt lm} (i.e.\ positive resp. negative links). To express a preference, such as $a > b$, we use the predicate {\tt pref(a,b)}. Note that DIAMOND provides a method to translate propositional formula ADFs and PADFs into ADFs with total functions and only computes the models using the functional representation.


For illustration, let us look at another, slightly more complicated example.

\begin{example}\label{exa:adf2}%
  Consider the ADF $D_2=(S_2,L_2,C_2)$ with $S_2=\{a,b,c,d\}$, $L_2=\{(a,c),(b,b),(b,c),(b,d)\}$, and $C_2=\{\varphi_a=\tvt, \varphi_b=b, \varphi_c=a \land b, \varphi_d=\lnot b\}$. 
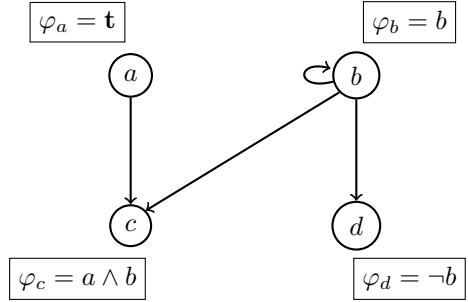
  \begin{center}\begin{tikzpicture}[->,thick,circle]
  \node (a) at (0,0) [draw] {$a$};
  \node (b) at (3,0) [draw] {$b$};
  \node (c) at (0,-2) [draw] {$c$};
  \node (d)  at (3,-2) [draw] {$d$};
  \node (va) [above left of=a,draw,thin,rectangle] {$\varphi_a=\tvt$};
  \node (vb) [above right of=b,draw,thin,rectangle] {$\varphi_b=b$};
  \node (vc) [below left of=c,draw,thin,rectangle] {$\varphi_c=a\wedge b$};
  \node (vd) [below right of=d,draw,thin,rectangle] {$\varphi_d=\neg b$};
  \path (a.south) edge[] (c.north);
  \path (b.south west) edge[] (c.north east);
  \path (b) edge[loop left] (b);
  \path (b.south) edge[] (d.north);
\end{tikzpicture}

\end{center}
For this ADF there are
\begin{itemize}
\item 16 admissible interpretations:
  $\emptyset$,
  $\{a\}$, $\{b\}$, $\{\lnot b\}$,
  $\{b,\lnot d\}$, $\{a,b\}$, $\{a,\lnot b\}$, $\{\lnot b,d\}$, $\{\lnot b,\lnot d\}$,
  $\{a,b,c\},$ $\{a,b,\lnot d\}$, $\{a,\lnot b,d\}$, $\{a,\lnot b,\lnot c\}$, $\{\lnot b,\lnot c,d\}$,
  $\{a,b,c,\lnot d\}$, $\{a,\lnot b,\lnot c,d\}$
\item three complete models:
  $\{a\}$, $\{a,b,c,\lnot d\}$, $\{a,\lnot b,\lnot c,d\}$; of these, $\{a\}$ is the grounded model;
\item two models:
  $\{a,b,c,\lnot d\}$, $\{a,\lnot b,\lnot c, d\}$, of which one is stable: $\{a,\lnot b,\lnot c,d\}$.
  Its propositional formula representation for DIAMOND (inherited from {\tt ADFsys}) is given by the following ASP code:
\newpage
\begin{slisting}
  \> statement(a). statement(b). statement(c). statement(d).\\
  \> ac(a,c(v)).\\
  \> ac(b, b).\\
  \> ac(c, and(a,b)).\\
  \> ac(d, neg(b)).
\end{slisting}
The functional ASP representation of the same ADF looks thus:
\begin{slisting}
  \> s(a). s(b). s(c). s(d).\\
  \> l(a,c). l(b,b). l(b,c). l(b,d).\\
  \> ci(a).\\
  \> co(b). ci(b,1,b).\\
  \> co(c). co(c,1,a). co(c,2,b). ci(c,3,a). ci(c,3,b).\\
  \> ci(d). co(d,1,b).
\end{slisting}
\end{itemize}
\end{example}

Arguably, the formula representation is easier to read for humans.

\presubsection
\subsection{Implementation of $\Gamma_D$}
\postsubsection

Since all of the semantics are defined via the operator $\Gamma_D$, we will now present how the implementation of the operator is done in DIAMOND. 
The unary predicate {\tt step} with an arbitrary term is used to apply the operator several times. 
The input for the operator is given by the predicates {\tt in} and {\tt out} to represent mappings to $\tvt$ and $\tvf$. 
The resulting interpretation can be read off the predicates {\tt valid} and {\tt unsat}.
Predicates {\tt fp} and {\tt nofp} denote whether a fixpoint is reached or not.
First, DIAMOND decides which of the mappings to $\tvt$ are still of interest ({\tt cii}) (i.e.\ which of those can still be satisfied under the given interpretation):
\begin{slisting}
  \> ciui(S,J,I) :- lin(X,S,I), not ci(S,J,X), ci(S,J). \\ 
  \> ciui(S,J,I) :- lout(X,S,I), ci(S,J,X).             \\
  \> cii(S,J,I) :- not ciui(S,J,I), ci(S,J), step(I).  
\end{slisting}
The predicates {\tt lin} and {\tt lout} are those links between arguments which are already decided by the given three-valued interpretation. The binary predicate {\tt ci} (resp. {\tt co}) is just the projection of its tertiary version to express that at least one predicate with a specific statement occurs in a specific acceptance condition.
The treatment of the interesting mappings to $\tvf$ ({\tt coi}) is dual:
\begin{slisting}
  \> coui(S,J,I) :- lin(X,S,I), not co(S,J,X), co(S,J). \\
  \> coui(S,J,I) :- lout(X,S,I), co(S,J,X).             \\
  \> coi(S,J,I) :- not coui(S,J,I), co(S,J), step(I).
\end{slisting}
Afterwards it is checked whether there exist two-valued extensions of the given interpretation that are a model or not, which is denoted by the predicates {\tt pmodel} (resp. {\tt imodel}). 
Then a statement can be seen to be {\tt valid} (resp.\ {\tt unsat}) if there does not exist an interpretation which is not a model (is a model)\/. The predicate {\tt verum} (resp. {\tt falsum}) represents that the acceptance condition is always true (resp.\ false).\newpage
\begin{slisting}
  \> pmodel(S,I) :- cii(S,J,I). pmodel(S,I) :- verum(S), step(I).\\
  \> pmodel(S,I) :- not lin(S,I), ci(S), step(I).\\
  \> pmodel(S,I) :- not lin(S,I), ci(S), step(I).\\
  \> valid(S,I) :- pmodel(S,I), not imodel(S,I).\\
  \> \\
  \> imodel(S,I) :- coi(S,J,I).\\
  \> imodel(S,I) :- falsum(S), step(I).\\
  \> imodel(S,I) :- not lin(S,I), co(S), step(I).\\
  \> unsat(S,I) :- imodel(S,I), not pmodel(S,I).
\end{slisting}
At last, either {\tt nofp} or {\tt fp} is deduced. 
To achieve this, DIAMOND checks whether the application of the operator resulted in an interpretation that is different from the given one.
\begin{slisting}
  \> nofp(I) :- in(X,I), not valid(X,I), step(I).\\
  \> nofp(I) :- valid(X,I), not in(X,I), step(I).\\
  \> nofp(I) :- out(X,I), not unsat(X,I), step(I).\\
  \> nofp(I) :- unsat(X,I), not out(X,I), step(I).\\
  \> fp(I) :- not nofp(I), step(I).
\end{slisting}

\presubsection
\subsection{Semantics}
\label{sec:semantics}
\postsubsection

The admissible model is computed by the use of a guess and check approach. 
At first a three-valued interpretation is guessed, by an assignment of the statements to be {\tt in}, {\tt out}, or neither. 
The last two lines remove all guesses which violate the definition of the admissible model (i.e.\ check which guesses are right)\/:
\begin{slisting}
\> step(0).\\
\> \{in(S,0):s(S)\}.\\
\> \{out(S,0):s(S)\}.\\
\> :- in(S,0), out(S,0).\\
\> :- in(S), not valid(S,0).\\
\> :- out(S), not unsat(S,0).
\end{slisting}

The complete model encoding uses the same concept as used for the admissible model. 
The only difference is that the guessed model needs to be a fixpoint.
To this effect the last two rules of the above encoding are replaced by the constraint ``{\tt :- nofp(0).}''.

To compute the grounded model, we need to apply $\Gamma_D$ until a fixpoint is reached. 
This is done via a sequence of steps, where the result of one step is taken as the used given interpretation for the next step\/:

\begin{slisting}
\> maxit(I) :- I:=\{s(S)\}. step(0). \\ 
\> in(S,I+1) :- valid(S,I). out(S,I+1) :- unsat(S,I).\\
\> step(I+1) :- step(I), not maxit(I).\\
\> in(S) :- fp(I), in(S,I).\\
\> out(S) :- fp(I), out(S,I).\\
\> udec(S) :- fp(I), s(S), not in(S), not out(S).
\end{slisting}
Note that we use the number of statements as the upper bound on the number of operator applications as this is the maximal number of steps needed to reach a fixpoint.

To implement the model semantics, the operator is not essential:
as the model is only two-valued, there do not remain undecided parts. 
So each variable is mapped to a truth-value and therefore every acceptance condition may only map to one value (i.e.\ $\tvt$ or $\tvf$). 
The encoding just guesses a two-valued interpretation and checks whether the guessed interpretation agrees with the acceptance conditions of each statement or not.
The stable model combines the encoding for models with the operator encoding to check for each model whether it is also the grounded extension of its reduced ADF or not. 

\presection
\section{Discussion and Future Work}
\label{sec:discussion}
\postsection

We presented the DIAMOND software system that uses answer set programming to compute models of abstract dialectical frameworks under various semantics.
DIAMOND can be seen as a continuation of the trend to utilise ASP for implementing abstract argumentation.
The most important existing tool in this line of work is the ASPARTIX system\footnote{ASPARTIX is available at \url{http://www.dbai.tuwien.ac.at/research/project/argumentation/systempage/}}~\cite{EglyGW10} for computing extensions of Dung argumentation frameworks. 

Quite recently, Ellmauthaler and Wallner presented their system {\tt ADFsys}\footnote{{\tt ADFsys} is available at \url{http://www.dbai.tuwien.ac.at/research/project/argumentation/adfsys/}} for determining the semantics of ADFs~\cite{Ellmauthaler2012}.
Since their system likewise uses answer set programming, it is natural to ask where the differences lie.
For one, after the discovery of several examples where some original ADF semantics do not behave as intended, Brewka et al.~\cite{brewka13adfs} proposed revised and generalised versions of these semantics.
The DIAMOND system implements the new semantics while {\tt ADFsys} still computes the old versions.
For another, {\tt ADFsys} relies solely on the representation of acceptance conditions via propositional formulas, while DIAMOND can additionally deal with functional representations.
Due to the new semantics it is not trivial to compare those two systems. 
In fact only the model and the grounded semantics have not changed. 
During preliminary tests, we used different methods to generate randomised ADF instances. 
Depending on the used generation method, DIAMOND could compete with {\tt ADFsys} and even outperform it. 
Alas, there were also instances for which {\tt ADFsys} outperformed DIAMOND. 
We consider it an important future task to determine specific classes of ADFs that distinguish the two systems, and to connect these ADF classes to possible real-world applications.

To adapt {\tt ADFsys} to the new semantics, it would be needed to decide at each operation of $\Gamma_D$ which acceptance formulae are (under the given three-valued interpretation) irrefutable (resp.\ unsatisfiable). To solve such an embedded co-\NP{} problem it would be necessary to use the saturation technique or similar concepts, which will make the use of disjunctive logic programs obligatory. Therefore there would also be issues with more complex semantics (like the preferred semantics). There the use of meta-ASP would conflict with the use of saturation in the disjunctive program.

Apart from the semantics implemented in this paper, there are also ADF semantics that DIAMOND cannot yet deal with -- these remain for future work.
For example, the preferred semantics is based on maximisation, and so we will need meta-ASP to implement that.
In general, ADFs are a quite new formalism, and we expect that further ADF semantics will be defined in the future.
Naturally, we plan to implement these new semantics using the infrastructure already available through DIAMOND.

Another future research interest concerns a possible practical application for ADFs:
We intend to analyse discussions in social media, where opinions and viewpoints can be modelled by statements that are in some relation to each other.
ADF semantics can guide the respective online community, for example as to what positions everybody can agree on, or how a group decision can be justified.
Such an approach was proposed by Toni and Torroni~\cite{Toni2011} as a possible application of assumption-based argumentation frameworks~\cite{BondarenkoDKT97}.
However, assumption-based argumentation inherits the expressiveness limitations of abstract argumentation, that is, it can also express only attack relationships between statements.
We expect that ADFs with their greater expressiveness are better suited to model online interactions in social media.

A similar application of argumentation in online social communities is the approach by Snaith~et~al.~\cite{Snaith2012}.
They utilise their database for arguments in the Argument Interchange Format~\cite{Rahwan2009} to capture discussions via different blogging-sites and use their tool TOAST~\cite{Snaith2012a} to compute an acceptable consensus about the issues under discussion. 
Again we think that ADFs are more suitable for this application due to their expressiveness.

\presubsection
\bibliographystyle{splncs}
\bibliography{paper}

\end{document}